# Domain-Adaptive Small Language Models for Structured Tax Code Prediction

**Souvik Nath Sumit Wadhwa Luis Perez**

Dell Technologies

souvik.nath1@dell.com, sumit_wadhwa@dell.com, luis.perez@dell.com

## Abstract

Every day, multinational firms process thousands of transactions, each of which must adhere to tax regulations that vary by jurisdiction and are often nuanced. The determination of product and service tax codes, such as HSN or SAC, is a major use case in tax compliance. An accurate determination of such codes is imperative to avoid any tax penalties. This paper proposes a domain-adaptive small language model (SLM) with an encoder-decoder architecture for the enhanced prediction of product and service tax codes. In this approach, we address the problem of predicting hierarchical tax code sequences using unstructured product and service data. We employ an SLM based on an encoder-decoder architecture, as this enables sequential generation of tax codes to capture the hierarchical dependencies present within the tax codes. Our experiments demonstrate that encoder-decoder SLMs can be successfully applied to the sequential prediction of structured tax codes, a domain that remains comparatively unexplored in current NLP research. In this paper, we demonstrate the superior performance of the domain-adaptive encoder-decoder SLMs over flat classifiers when applied to the Harmonized System of Nomenclature (HSN), and achieve superior results compared to decoder-only and encoder-only architectures for structured sequence generation tasks. This approach can also be scaled to other government-mandated tax commodity codes, such as United Nations Standard Products and Services Codes (UNSPSC), or Brazil's Nomenclatura Comum do Mercosul (NCM).

## I. INTRODUCTION

Product and service tax code prediction plays a pivotal role in international trade, tax legislation, and supply chain management. The Harmonized System (HS) [1], [2] is widely regarded as the gold standard for classification of products and services in the context of taxation and customs. Inaccuracies in code assignments lead to financial discrepancies, compliance issues, and logistical inefficiencies.

Traditional methods such as rule-based systems or flat label classifiers are widely used even today in tax determination processes. These methods treat tax codes as whole entities, ignoring any structure within them. The hierarchical structure of these codes holds meaningful information and plays a significant role in their determination. The HS, maintained by the World Customs Organization, classifies goods with a 6-digit code built from three nested 2-digit levels: the chapter (first two digits), the heading (next two) and the subheading (final two). India's Harmonized System of Nomenclature (HSN) extends this to 8 digits by appending a 2-digit national tariff item. Services are classified separately under the 6-digit Services Accounting Code (SAC), in which every code begins with chapter 99, followed by a 2-digit heading, a 1-digit group and a 1-digit service code. Generally, these traditional methods are non-sequential and fail to capture and learn from inherent hierarchical structures present in these tax codes, thereby overlooking critical signals and suffering from reduced accuracy and poor interpretability. To address this, we framed our task as a structured sequence prediction problem, breaking the tax code into its hierarchical components (chapter, heading, subheading, and tariff) and predicting them one by one.

With recent advancements in pre-trained language models, we can fine-tune small language models (SLMs) [3] on domain-specific data on cost-effective infrastructure. This domain adaptation of SLMs enables strong alignment with the structured nature of taxonomies like HSN or SAC, foregoing the need for complex rule-based systems or massively scaled models. This aligns with recent findings [4] showing that SLMs offer significant gains in cost and scalability. Building on this formulation, we propose domain-adaptive small language models to predict a structured sequence of tax codes. By fine-tuning a pre-trained encoder-decoder model on domain-specific data related to taxation, the model learns to generate tax code components in alignment with the taxonomy of HSN or SAC. This is analogous to how alignment is achieved in neural machine translation systems [5]. As demonstrated in this paper, the proposed approach significantly improves prediction accuracy and enhances model explainability by aligning the model's generation process with the underlying taxonomy of the codes.

## II. RELATED WORK

### A. AUTOMATED HS CODE CLASSIFICATION

Automated assignment of HS codes from product descriptions has been studied as both a retrieval and a supervised classification problem. Spichakova and Haav [6] assessed the correctness of declared HS codes by combining Doc2Vec text similarity with taxonomy-based semantic similarity across HS levels. Stassin et al. [7] compared similarity-based ranking using pre-trained sentence embeddings against supervised neural classifiers for HS6, HS8, and HS10 prediction, and Lee et al. [8] proposed an explainable product classifier for customs that retrieves supporting text from the HS manual. These approaches predict codes as flat labels or rank candidate descriptions rather than generating the code level by level, and they target customs declarations rather than enterprise transaction-tax determination for goods and services.

### B. HIERARCHICAL CLASSIFICATION AS SEQUENCE GENERATION

Hierarchical classification methods are commonly grouped into flat, local, and global approaches [9]. Global models encode the label taxonomy directly, for example through structure encoders over the label hierarchy [10]. A separate line of work casts label prediction as sequence generation with encoder–decoder models [11], [12], and constrains decoding so that generated labels remain consistent with the hierarchy [13]. Constrained decoding over valid outputs has likewise been used to guarantee valid identifiers in autoregressive entity retrieval [14], [15]. Our work brings these ideas to tax code prediction: we serialize HSN and SAC codes into taxonomy-aware token sequences, fine-tune a small encoder–decoder model, and restrict decoding to valid codes at each level, comparing against encoder-only, decoder-only, and flat baselines on expert-labelled enterprise data.

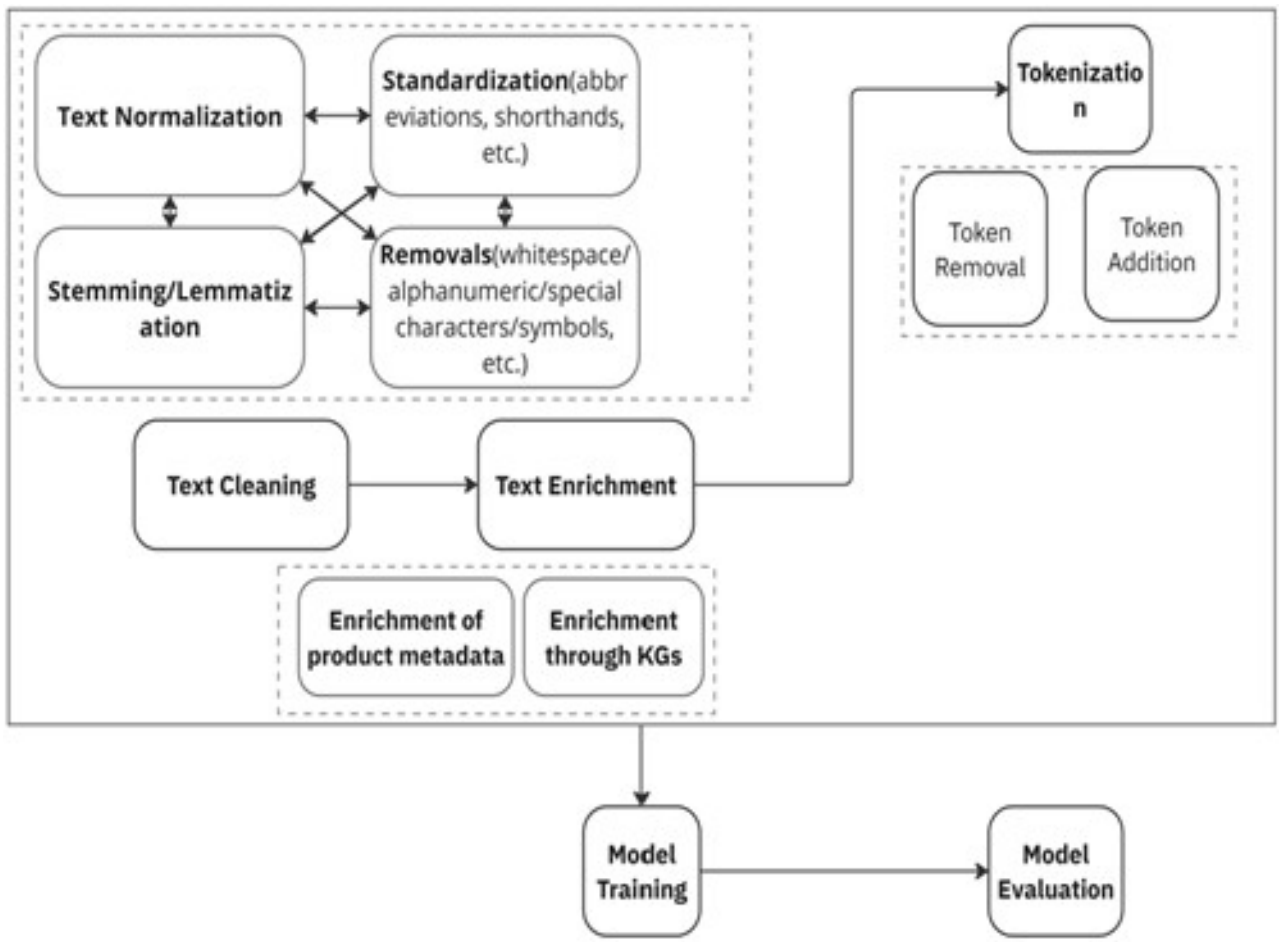

**FIGURE 1. Data processing workflow for handling unstructured text data**

## III. METHODOLOGY

Natural language processing (NLP) solutions are structured around key sequential stages, including feature collection, data cleaning (tokenization and stemming), text normalization, data enrichment, model training, and evaluation, as depicted in Figure 1. We provide more details in the upcoming sections covering details specific to the given problem statement.

### A. TEXT CLEANING

Text cleaning is a critical step in natural language processing tasks. It transforms raw textual data into a structured format that is suitable for analysis and modelling. Generally, various stages of preprocessing are designed to remove inconsistencies, irrelevant information, and noise from the data.

The following steps were taken to ensure data consistency and integrity. De-duplication of text: duplicate phrases within product descriptions were identified and removed. We detected and eliminated repeated sub-sequences for each product description. This reduction in redundancy sets the text up for further processing. Alphanumeric and special token removal: to maintain relevance and reduce noise in the dataset, alphanumeric tokens, including serial numbers, batch identifiers and irrelevant special characters, were filtered out. Tokens determined to be irrelevant based on contextual analysis were removed, preserving only information meaningful for classification. Text normalization and standardization are essential to reduce variability in textual data. In our approach, we treated different variations of product names such as a 2-in-1 laptop or tablet appearing as “two in one,” “2in1,” etc., by standardizing them to a common representation. We treated inconsistent text formatting as non-informative noise and standardized it to improve the quality of embeddings. This was done to avoid any artificial distinction between otherwise similar product descriptions, and the step concluded with lemmatization. Specialized token handling: edge cases for specific phrases and sub-phrases were handled with the objective of removing bias. Specific tokens with no relevance in the semantic context of product descriptions were removed to avoid false associations, ensuring that only relevant data is retained. Brand information was masked to prevent bias towards brand names. The final refinement was to identify and remove incomplete product descriptions. Records containing ambiguous or incomplete information post-cleaning were removed to ensure the quality and reliability of the data fed into subsequent stages. This comprehensive approach to text cleaning significantly improved data quality, enabling a robust and effective foundation for further NLP processing and tax code classification.

### B. TEXT ENRICHMENT

Text enrichment is an essential stage in natural language processing. It is designed to enhance the contextual and semantic richness of textual data. In text enrichment, we include additional relevant data sourced externally from databases, knowledge graphs, or even domain-specific structured corpora such as operational manuals. This process improves the descriptive quality and comprehensiveness of the text, which enhances the accuracy of subsequent modelling tasks.

We enhanced product and service descriptions by integrating product specifications and metadata (not present in the description) fetched from an internal product database.

Products are often characterized by unique alphanumeric codes, and these codes do not contribute to the semantic context in a product description. Therefore, they are either standardized, expanded, or removed; for example, abbreviations were expanded to their full forms. We employed matching (similarity) techniques [16] to find similar product descriptions based on the product type. We used product categorization information, such as form factor and portability, to identify standardized product descriptions and add them to existing data for further enrichment. These steps elevated the semantic quality and contextual relevance of product descriptions, which contributed to improved performance and generalization of the model.

### C. TOKENIZATION

Tokenization is a critical pre-processing step for training deep neural language models for natural language generation tasks. In this step, raw text is broken down into smaller units called tokens. Generally, tokens can be words, subwords, or characters, depending on the model and its use case. The choice of a tokenization strategy depends upon downstream tasks.

Each model was used with its native pre-trained tokenizer: SentencePiece [17] for T5, byte-level Byte-Pair Encoding (BPE) [18] for DistilGPT2, and WordPiece [19] for BERT. These subword schemes manage a diverse vocabulary and mitigate issues related to rare or out-of-vocabulary tokens. Each stock tokenizer was extended with domain-specific vocabulary derived from the unique terminology in our dataset, enabling it to capture industry-specific terms and product and service identifiers. We also added specialized taxonomy tokens (e.g., <hsn_ch_84>), whose role is discussed in Section III-D.

### D. MODEL DEVELOPMENT

#### *1) DATA AND MODEL SETUP*

We fine-tuned the T5-small checkpoint (60.5M parameters) on a proprietary corpus of approximately 100,000 unique product and service descriptions, obtained after de-duplication of a substantially larger pool of historical transaction records. Codes with fewer than 100 training records were merged into a single low-volume class, yielding a few hundred distinct HSN/SAC classes. The corpus was split 80/20 into training and validation sets, and all models were evaluated on a separate expert-validated test set of 16,000 records. Training and inference engines were developed on PyTorch [20]. We initialized our models with the weights of pre-trained models from the Hugging Face ecosystem [21] and performed fully supervised fine-tuning on tax-domain data, enabling domain-specific adaptation. To facilitate scalability, we implemented data parallelism, allowing rapid experimentation. Training parameters such as batch size and learning rate were empirically optimized to ensure performance stability during training, and an adaptive learning rate schedule was employed to improve convergence.

#### *2) MODEL SELECTION AND TRAINING*

Transformer-based language models [22] represent the state-of-the-art in natural language processing. Leveraging such pretrained SLMs and post-training them to adapt to domain-specific tasks has been shown to be both cost-effective and time-efficient [4].

SLMs can be designed using three primary architectures: encoder-only, encoder-decoder and decoder-only. In this work, we conducted experiments using pre-trained language models from each of these architectures. We utilized compact models and associated techniques referenced in prior work [23], [19], [24].

Our main goal was to map free-form product descriptions to hierarchical tax codes such as HSN/SAC. This task presents a major challenge due to the asymmetry between free-form input text and structured hierarchical output, something we normally see in cross-domain mapping problems.

Initial results revealed that the T5 model, based on an encoder-decoder architecture, consistently outperformed the decoder-only (DistilGPT2 [25]) and encoder-only (BERT) models. This superior performance can be attributed to the inherent advantages of the encoder-decoder architecture: its ability to generate richer input representations, its clear separation of input and output processing, and its effectiveness in cross-space mapping tasks.

Our task of determining the appropriate hierarchical tax codes given product descriptions is analogous to neural machine translation, where unstructured textual inputs are mapped to an ordered sequence of categorical codes. Prior works [26], [5], [27], [28] have shown the effectiveness of using such architectures for cross-domain mapping tasks. All evaluated models, including encoder-only, decoder-only, and encoder-decoder architectures, were fine-tuned on our domain-specific data. Among these, the T5 encoder-decoder model showed strong performance, effectively capturing both lexical semantics and hierarchical dependencies present in the tax code structure. Fine-tuning further improved its generalization capability, reinforcing its suitability for structured sequence prediction tasks such as hierarchical tax code assignment.

Given the strong performance of T5 in this structured prediction setting, we next describe how its output generation process was adapted to align with the hierarchical tax codes. The target output sequence was structured in a stepwise manner, retaining the taxonomy of these codes by decomposing them into four stages. In Figure 2, we show how an HSN/SAC code can be decomposed into smaller entities based on their taxonomy.

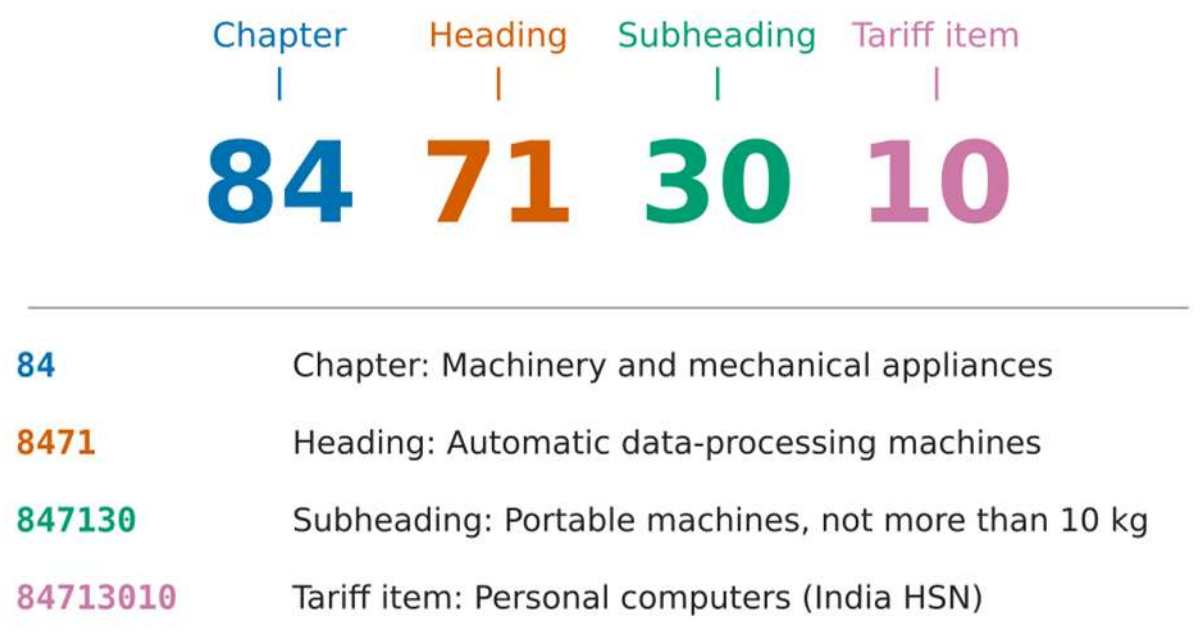

**FIGURE 2. Hierarchical structure of an 8-digit HSN code (example: 84713010)**

1) **Chapter Selection:** Chapter is at the first position of the output sequence. The model first selects the best Chapter based on the highest probability $P(Ch_i \mid X)$. This ensures that the selected chapter is contextually relevant to the product description.

2) **Heading Selection:** Every Chapter has a unique set of Headings. During text generation, the model constrains the candidate list to only those Headings that belong to the selected Chapter and selects the one with the highest conditional probability $P(He_i \mid X, Ch_i)$.

3) **Sub-Heading Selection:** For the chosen Chapter and Heading, there is again a unique set of Sub-Headings; the model selects the Sub-Heading with the highest conditional probability $P(SH_i \mid X, Ch_i, He_i)$.

4) **Product Tariff Selection:** Finally, the Product Tariff is selected from the relevant list based on the choices made in the previous time steps. The Product Tariff with the highest conditional probability $P(T_i \mid X, Ch_i, He_i, SH_i)$ is selected. For SAC codes, the same procedure applies, with the group and service code forming the third and fourth levels. For the generative models, decoding is restricted at each level to codes that are valid under the selected parent, so that every prediction is a valid tax code; the flat classifiers satisfy the same requirement by construction.

We introduced a set of special tokens to aid these models in learning local semantics within the tax codes. For example, an 8-digit HSN code such as 84713010 is decomposed into the sequence <hsn_ch_84> <hsn_h_71> <hsn_sh_30> <hsn_pt_10>. A 6-digit SAC code follows the official SAC hierarchy of chapter, heading, group and service code; for example, 997111 is decomposed into <sac_ch_99> <sac_h_71> <sac_gr_1> <sac_sc_1>. Serializing the hierarchical tax codes into a flat output sequence with special tokens marking the code type ('hsn'/'sac') and level ('ch', 'h', 'sh', 'pt' for HSN; 'ch', 'h', 'gr', 'sc' for SAC) allowed the model to learn distinct embedding representations for each structural unit. Eventually, this helped the model to differentiate between code segments and further facilitated the learning of local dependencies between adjacent levels, thus ensuring global alignment with the overall underlying taxonomy of these tax codes.

During early experiments, we observed a bias towards overused placeholder tokens such as ***<UNK>.*** These tokens appeared in place of missing product names, either due to their absence in the training data or inconsistencies in the source description. Inducing noise and perturbations to the data during training helped overcome this bias, leading to performance improvements. We also applied label smoothing to prevent overconfidence and improve generalization across tax codes. These steps contributed to the domain adaptation by ensuring better generalization across real-world tax prediction scenarios.

As discussed so far, all the strategies considered were aimed at reinforcing the model's alignment with domain-specific structures in our data. Instead of treating this as a generic sequence generation or classification task, each component of our training pipeline was specifically designed to preserve semantics in hierarchical dependencies of these tax codes. Such fine-grained domain adaptation, based on a lightweight encoder-decoder architecture, played a key role in enabling robust generalization of the model across a diverse set of products and tax codes.

## IV. RESULTS, ABLATION, AND DEPLOYMENT ANALYSIS

### A. MODEL SCOPE AND EVALUATION SETUP

In this study, we focus on Small Language Models (SLMs), with particular emphasis on encoder–decoder sequence-to-sequence architectures, which are inherently well suited for generating structured and hierarchical outputs. All SLMs considered in this work were fine-tuned on domain-specific tax data, enabling them to learn the hierarchical dependencies present in HSN and SAC tax codes.

To assess the impact of architectural choice, we conducted a systematic comparison across encoder–decoder (T5), encoder-only (BERT), and decoder-only (DistilGPT-2) models, alongside a traditional flat Multi-Layer Perceptron (MLP) baseline. T5 and DistilGPT-2 generate the code as a hierarchical token sequence, whereas BERT and the MLP are trained as flat classifiers in which each complete HSN or SAC code is treated as a single class. BERT uses a softmax classification head over its [CLS] representation. The MLP takes term frequency–inverse document frequency (TF-IDF) features of the same preprocessed product descriptions as input and uses a compact fully connected architecture, which accounts for its small parameter count (Table II).

All models were evaluated on a high-quality test set of 16,000 records, with HSN/SAC codes assigned and validated by Tax and Procurement experts, ensuring a reliable and expert-grounded performance comparison. A prediction is counted as correct only if the complete code (all 8 HSN or 6 SAC digits) exactly matches the expert-assigned code; partial matches at higher levels of the hierarchy receive no credit. Precision, recall, and F1-score are macro-averaged across codes, so that performance on low-volume codes is weighted equally with high-volume codes. All models evaluated in this study, including encoder-only, decoder-only, and encoder–decoder architectures, were trained and evaluated using a common input data preprocessing, normalization, and enrichment pipeline, enabling a controlled comparison across architectures. Differences in target representation and output generation were intentionally aligned with the capabilities of each architectural paradigm.

### B. COMPARATIVE PERFORMANCE ACROSS ARCHITECTURES

Table I summarizes performance across the evaluated fine-tuned SLMs. Among all architectures considered, the encoder–decoder T5 model consistently outperformed encoder-only and decoder-only alternatives, achieving the highest precision (70%) and F1-score (65%). Encoder-only (BERT) and decoder-only (DistilGPT-2) models exhibited moderate performance, while the flat Multi-Layer Perceptron (MLP) baseline failed to capture the hierarchical structure inherent in HSN and SAC codes.

Overall, these results indicate that architectural suitability for structured sequence generation, combined with supervised domain-specific fine-tuning, plays a decisive role in hierarchical tax code prediction, favoring encoder–decoder SLMs for this class of tasks.

TABLE I
COMPARISON OF ACCURACY MEASURES

| Metric | T5 | DistilGPT 2 | BERT | MLP |
|---|---|---|---|---|
| **Precision** | 70% | 66% | 55% | 20% |
| **Recall** | 61% | 60% | 58% | 23% |
| **F1-Score** | 65% | 62% | 56% | 22% |

### C. TEMPORAL AGREEMENT WITH DOMAIN EXPERTS

To assess robustness over time, we computed Cohen's kappa [29] between model predictions and expert annotations over an 8-month period. Cohen's kappa values reflect agreement beyond chance, where 1.0 denotes perfect agreement.

As shown in Figure 3, T5 achieved the highest mean monthly agreement ($\kappa \approx 0.56$), followed by DistilGPT-2 ($\approx 0.48$), BERT ($\approx 0.43$), and MLP ($\approx 0.09$). Agreement for the fine-tuned models was lower in the first two months and improved as the evaluation progressed, with a modest dip in months 6–7 that we attribute to data drift. T5 achieved the highest or joint-highest agreement in seven of the eight months, reinforcing its suitability for real-world tax determination.

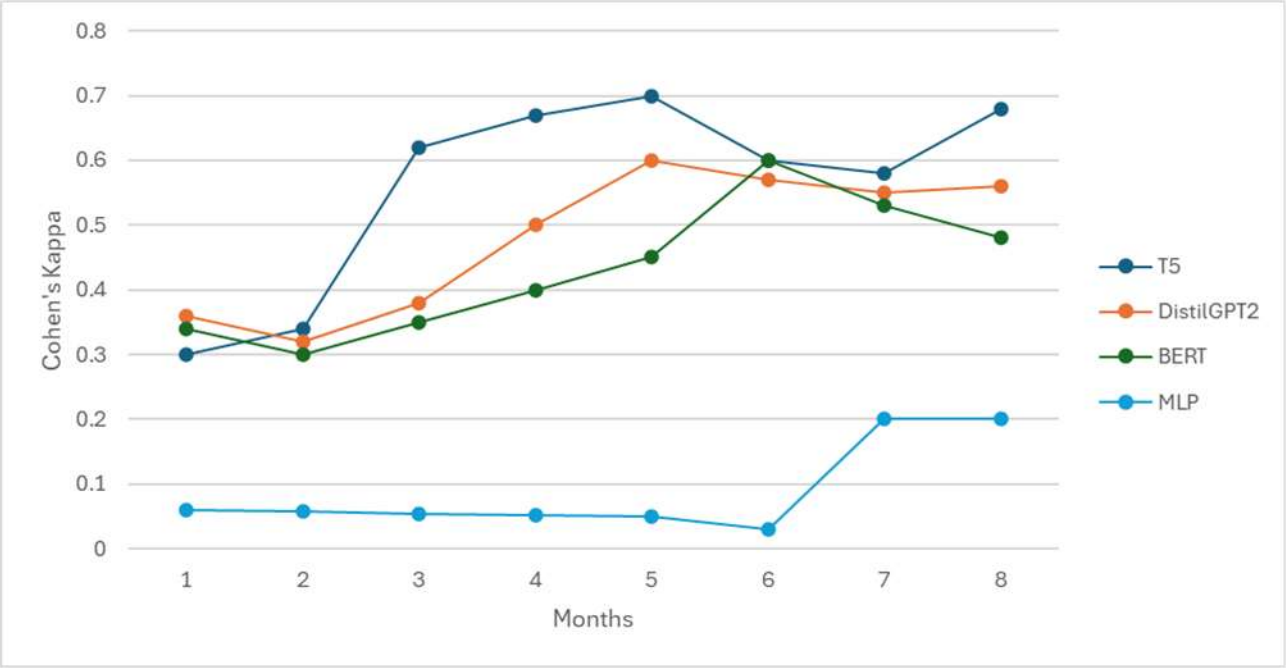

**FIGURE 3.** **Cohen's Kappa(*k*) over time**

### D. ABLATION STUDY

To rigorously examine the sources of performance differences, we conducted a multi-stage ablation study designed to disentangle the effects of model architecture from output representation strategy.

#### 1) ARCHITECTURAL COMPARISON

We first compared three SLM architectural paradigms—encoder-only (BERT), decoder-only (DistilGPT-2), and encoder–decoder (T5)—all fine-tuned on the same domain-specific dataset. This experiment isolates the model's ability to map free-form textual descriptions to hierarchically structured output spaces, a setting characterized by strong input–output asymmetry. Encoder-decoder models consistently outperformed encoder-only and decoder-only alternatives, confirming their suitability for structured sequence generation tasks such as tax code assignment.

TABLE II
COMPARISON OF MODEL SIZE
BY NUMBER OF PARAMETERS

| Model | # Params |
|---|---|
| **T5** | 60.5M |
| **DistilGPT2** | 88.2M |
| **BERT** | 110M |
| **MLP** | 105k |

*2) STRUCTURED VS. UNSTRUCTURED TARGET GENERATION*

In a second ablation stage, we evaluated the impact of explicitly structuring the target output sequence using domain-specific taxonomic tokens. For both DistilGPT-2 and T5, we compared models trained to generate flat numeric codes against models trained to generate decomposed hierarchical sequences aligned with HSN/SAC taxonomy.

To isolate this effect, we ablated the use of our custom hierarchical tokenization, forcing the models to generate standard numeric sequences without special taxonomic markers (e.g., <hsn_ch_xx>). Reintroducing structured target tokens yielded consistent but secondary precision gains of 3–4 percentage points (DistilGPT-2: 62% to 66%; T5: 67% to 70%). The gain was larger for the decoder-only model, which lacks a dedicated encoder and benefits more from explicit structural markers. Notably, T5 without structured tokens already outperformed DistilGPT-2 with them, confirming that the encoder–decoder architecture is the primary driver of performance.

These results indicate that while architectural choice remains the primary driver of performance, explicitly aligning the generation process with the underlying taxonomy further amplifies accuracy and stability in hierarchical tax code prediction.

TABLE III
PERFORMANCE COMPARISON AND ABLATION RESULTS

| **MODEL** | **PRECISION** | **RECALL** |
|---|---|---|
| *Multi-Layer Perceptron* | *20%* | *23%* |
| BERT | 55% | 58% |
| DistilGPT2 w/o special tokens in target sequence | 62% | 60% |
| DistilGPT2 w/ special tokens in target sequence | 66% | 60% |
| T5 w/o special tokens in target sequence | 67% | 61% |
| T5 w/ special tokens in target sequence | 70% | 61% |

### E. ENTERPRISE SCALABILITY AND COST EFFICIENCY

Recent industry benchmarks [30] highlight a significant divergence in operational costs and latency when comparing Small Language Models (SLMs) with commercial Large Language Models (LLMs). While predictive accuracy is critical for avoiding compliance penalties, deploying natural language processing systems in enterprise environments also demands a highly scalable and cost-effective infrastructure.

To evaluate the scalability of our proposed domain-adaptive SLM approach, we conducted an enterprise-scale cost and computational efficiency analysis comparing an on-premises SLM deployment against commercial LLMs accessed via an application programming interface (API). We modelled an illustrative enterprise workload of multi-million invoices processed annually, assuming an average of 50 tokens per invoice, including both the input product description and the generated hierarchical tax code sequence. This corresponds to an annual processing volume of approximately 75–80 million tokens, or 6–7 million tokens per month.

These volumes reflect enterprise-scale inference across all transactions, including recurring product descriptions, and are distinct from the de-duplicated training corpus described in Section III-D. For a like-for-like comparison, both estimates assume the same token volume; this is conservative for the LLM, since prompting a general-purpose model typically requires additional instructions and examples. SLM costs are infrastructure-equivalent marginal estimates derived from [30] and exclude fixed hosting overhead. Beyond cost, on-premises SLM deployment keeps sensitive tax data within the enterprise, provides predictable low latency, and, through restricted decoding, guarantees structurally valid codes.

Recent industry benchmarks further indicate a stark contrast in both latency and operational expenditure between SLMs and LLMs [30]. According to enterprise decision guides, commercial LLMs accessed via API (such as GPT-4o) are associated with an estimated cost of $4.00–$5.00 per million tokens [30]. In contrast, an SLM incurs a baseline cost of approximately $0.04 per million tokens, and self-hosting compact models such as T5 (60.5M parameters) results in a cost-efficiency gap of 100× or more in favor of SLM-based deployments [30].

TABLE IV
COMPARISON OF ENTERPRISE EFFICIENCY AND COST

| Metric | LLM | SLM |
|---|---|---|
| **Estimated Latency (Per Request)** | 200-300 ms | 10-50 ms |
| **Estimated Cost (Per 1M Tokens)** | $4-$5 | $0.04(Infrastructure Equivalent) |
| **Estimated Monthly Compute Cost** | $31.25 | $0.25 |

## V. CONCLUSION

In this work, we addressed the challenge of mapping unstructured product and service descriptions to hierarchically structured tax codes such as HSN and SAC. We demonstrated that domain-adaptive **Small Language Models (SLMs)** based on **encoder–decoder architectures** are particularly effective for this task, outperforming encoder-only, decoder-only, and flat classification baselines when trained with supervised domain data.

By framing tax code determination as a structured sequence generation problem and aligning the model output with the underlying taxonomy, our approach preserves hierarchical dependencies and improves interpretability. While architectural choice emerged as the primary driver of performance, explicitly structuring target sequences further enhanced accuracy and stability.

In addition to predictive performance, we showed that compact, fine-tuned SLMs offer substantial advantages in cost and scalability compared to commercial LLMs accessed via API, making them well suited for large-scale enterprise tax compliance deployments. The proposed domain-adaptive encoder–decoder approach is not limited to HSN and SAC and can be extended to other regulated, taxonomy-driven classification systems, which represents a natural direction for future work.

## ACKNOWLEDGMENT

The authors thank the Intelligent Tax Systems Engineering team for their insightful feedback on the deployment of the encoder-decoder architecture. The authors also thank the tax and procurement experts at Dell Technologies who assisted in carefully curating the 16,000 test-set records used for evaluating the models. The authors acknowledge the use of generative AI assistants for language refinement in this paper.